\documentclass[11pt,a4paper]{article}
\usepackage[hyperref]{emnlp2020}
\usepackage{times}
\usepackage{latexsym}

\usepackage{CJKutf8}
\usepackage{color}

\usepackage{booktabs}
\usepackage[normalem]{ulem} 
\usepackage{graphicx}
\usepackage{float}
\usepackage{subcaption}
\usepackage{mdframed} %
\usepackage{arydshln}

\usepackage{multirow}

\usepackage{soul}
\usepackage{microtype}

\aclfinalcopy %
\def\aclpaperid{2559} %

\newcommand{\added}[1]{\textcolor{black}{#1}}

\newcommand{\fz}[1]{\footnotesize{#1}}

\title{OCNLI: Original Chinese Natural Language Inference}

\author{
	Hai Hu$^\dagger{}^\diamond$ ~
	Kyle Richardson$^\ddagger$ ~ 
	Liang Xu$^\diamond$ \\
	\textbf{Lu Li}$^\diamond$ ~
	\textbf{Sandra K\"{u}bler}$^\dagger$  ~
	\textbf{Lawrence S. Moss}$^\dagger$ \\
	$^\dagger$Indiana University, Bloomington, IN, USA\\
	$^\ddagger$Allen Institute for Artificial Intelligence, Seattle, WA, USA\\
	$^\diamond$Chinese Language Understanding Evaluation (CLUE) benchmark \\
	{\small {\tt \{huhai,skuebler,lmoss\}@indiana.edu};	{\tt kyler@allenai.org};} \\
	{\small {\tt xuliang@i-i.ai}; {\tt ll000@mails.ccnu.edu.cn}; {\tt CLUE@CLUEbenchmarks.com} }
}

\date{}

\begin{document}
\maketitle
\begin{abstract}

Despite the tremendous recent progress on natural language inference (NLI), driven largely by large-scale investment in new datasets (e.g., SNLI, MNLI) and advances in modeling, most progress has been limited to English due to a lack of reliable datasets for most of the world's languages. In this paper, we present the first large-scale NLI dataset (consisting of $\sim$56,000 annotated sentence pairs)\footnote{Our dataset and code are available at \url{https://github.com/CLUEbenchmark/OCNLI}.} for Chinese called the \textit{Original Chinese Natural Language Inference} dataset (OCNLI). Unlike recent attempts at extending NLI to other languages, our dataset does not rely on any automatic translation or non-expert annotation. %
Instead, we elicit annotations from native speakers specializing in linguistics. We follow closely the annotation protocol used for MNLI, but  create new strategies for eliciting diverse hypotheses. We establish several baseline results on our dataset using state-of-the-art pre-trained models for Chinese, and find even the best performing models to be far outpaced by %
human performance ($\sim$12\% absolute performance gap), making it a challenging new resource that we hope will help to accelerate progress in Chinese natural language understanding. \added{To the best of our knowledge, this is the first human-elicited MNLI-style corpus for a non-English language. }

\end{abstract}

\section{Introduction}
\label{sec:intro}

In the last few years, natural language understanding has made considerable progress,  driven largely by the availability of large-scale datasets and advances in neural modeling \cite{elmo,bert}. At the center of this progress has been natural language inference (NLI), which focuses on the problem of deciding 
whether two statements are connected via an entailment or a contradiction. 
NLI profited immensely from new datasets such as the  Stanford NLI (SNLI, \citet{snli}) and Multi-Genre NLI (MNLI, \citet{mnli}) datasets. %
However, as often the case, this progress has centered around the English language given that the most well-known datasets 
are limited to English. %
Efforts to build comparable datasets for other languages have largely focused on (automatically) translating existing English NLI datasets \cite{mehdad2011,xnli}. But this approach comes with its own issues (see section~\ref{sec:related}).

{ \renewcommand{\arraystretch}{1.5} %
\begin{CJK}{UTF8}{gbsn}
\begin{table*}[t]
{\footnotesize
\begin{tabular}{p{7cm}p{1.3cm}p{2cm}p{4cm}} \toprule
Premise & Genre\newline \texttt{Level} & \textbf{Majority label} \newline All labels & Hypothesis \\ \midrule
但是不光是中国，日本，整个东亚文化都有这个特点就是被权力影响很深 \newline But not only China and Japan, the entire East Asian culture has this feature, that is it is deeply influenced by the power. & TV\newline \texttt{medium} & \textbf{Entailment}\newline E E E E E & 有超过两个东亚国家有这个特点\newline  More than two East Asian countries have this feature. \\
完善加工贸易政策体\newline  (We need to) perfect our work and trade policies. & GOV\newline \texttt{easy} & \textbf{Entailment}\newline E E E E E & 贸易政策体系还有不足之处\newline  (Our) trade policies still need to be improved. \\
咖啡馆里面对面坐的年轻男女也是上一代的故事，她已是过来人了\newline  Stories of young couples sitting face-to-face in a cafe is already something from the last generation. She has gone through all that. & LIT\newline \texttt{medium} & \textbf{Contradiction}\newline C C C N N & 男人和女人是背对背坐着的\newline  The man and the woman are sitting back-to-back. \\
今天，这一受人关注的会议终于在波恩举行\newline  Today, this conference which has drawn much attention finally took place in Bonn. & NEWS\newline \texttt{easy} & \textbf{Neutral}\newline N N N N C & 这一会议原定于昨天举行\newline  This conferences was scheduled to be held yesterday. \\
嗯,今天星期六我们这儿,嗯哼.\newline  En, it's Saturday today in our place, yeah. & PHONE\newline \texttt{hard} & \textbf{Contradiction}\newline C C C C C & 昨天是星期天\newline  It was Sunday yesterday. \\ \bottomrule
\end{tabular}
}
\caption{Examples from the \textsc{MultiConstraint} elicitation of our Chinese NLI dataset, one from each of the five text genres. 
\texttt{easy}: 1st hypothesis the annotator wrote for that particular premise and label; \texttt{medium}: 2nd hypothesis; \texttt{hard}: 3rd hypothesis. \textbf{Bold} label shows the majority vote from the annotators. \label{tab:ocnli:examples} }
\end{table*}
\end{CJK}
} %

To overcome these shortcomings and contribute to ongoing progress in  Chinese NLU, we present the first large-scale NLI dataset for Chinese called the \emph{Original Chinese Natural Language Inference} dataset (OCNLI). Unlike previous approaches, we rely entirely on original Chinese sources and use native speakers of Chinese with special expertise in linguistics and language studies for creating hypotheses and for annotation. 
Our %
dataset contains $\sim$56,000 annotated premise-hypothesis pairs and follows a similar procedure of data collection to the English MNLI. Following MNLI, the premises in these sentence pairs are drawn from multiple genres (5 in total), including both written and spoken Chinese (see Table~\ref{tab:ocnli:examples} for examples). To ensure annotation quality and consistency, we closely mimic MNLI's original annotation protocols for monitoring annotator performance. We find that our trained annotators have high agreement on label prediction 
(with $\sim$98\% agreement based on a 3-vote consensus). To our knowledge, this dataset constitutes the first large-scale NLI dataset for Chinese that does not rely on automatic translation.

Additionally, we establish baseline results based on a standard set of NLI models \cite{chen2017lstm} tailored to Chinese, as well as new pre-trained Chinese transformer models \cite{cui2019pretraining}. 
We find that our strongest model, based on RoBERTa \cite{roberta}, performs far behind expert human performance ($\sim$78\% vs.\ %
$\sim$90\%  accuracy on our test data).
These results show that the dataset is challenging without using special filtering that has accompanied many recent NLI datasets \cite{le2020adversarial}. 

\paragraph{Contributions of this paper:} 1) We introduce a new, high quality dataset for NLI for Chinese, based on Chinese data sources and expert annotators; %
2) We provide strong baseline models for the task, and establish the difficulty of our task through experiments with recent pre-trained transformers. 3) We also demonstrate the 
benefit of naturally annotated NLI data by comparing performance with large-scale automatically translated datasets.

\section{Related Work}\label{sec:related}

Natural language inference (NLI), or recognizing textual entailment (RTE), is a long-standing task in NLP. Since we cannot cover the whole field, we focus on existing datasets and current systems.

\paragraph{Data:} To date, there exists numerous datasets for English, ranging from smaller/more linguistics oriented resources such as FraCaS \citep{fracas}, to larger ones like the RTE challenges \citep{Dagan} and SICK \citep{SICK}. Perhaps the most influential are the two large-scale, human-elicited datasets: the Stanford Natural Language Inference Corpus (SNLI) \citep{snli}, whose premises are taken from image captions, and the Multi-Genre Natural Language Inference Corpus (MNLI) \citep{mnli}, whose premises are from texts in 10 different genres. Both are built by collecting premises from pre-defined text, then having annotators come up with possible hypotheses and inference labels, which is the procedure we also employ in our work.

These large corpora have been used as part of larger benchmark sets, e.g., GLUE \citep{glue}, and have proven useful for problems beyond NLI, such as  sentence representation and transfer learning \citep{infersent,subramanian2018learning,reimers2019sentence}, automated question-answering \citep{khot2018scitail,trivedi2019repurposing} and model probing \cite{warstadt2019investigating,probing2020,geiger2020modular,jeretic2020natural}. %

\added{The most recent English corpus Adversarial NLI \citep{anli} uses Human-And-Model-in-the-Loop Enabled Training (HAMLET) method for data collection. Their annotation method requires an existing NLI corpus to train the model during annotation, which is not possible for Chinese at the moment, as there exists no high-quality Chinese data.    }

\added{In fact,} there has been relatively little work on developing large-scale human-annotated resources for languages other than English. Some NLI datasets exist in other languages, e.g., \citet{assin1} and \citet{assin2} for Portuguese,   \citet{hayashibe2020japanese} for Japanese, and \citet{amirkhani2020farstail} for Persian, but none of them have human elicited sentence pairs. 
Efforts have largely focused on automatic translation of existing English resources \cite{mehdad2011},  sometimes coupled with smaller-scale hand annotation by native speakers \cite{negri2011divide,agic2017baselines}. This is also true for some of the datasets included in the recent Chinese NLU benchmark CLUE \cite{clue} and for XNLI  \cite{xnli}, a multilingual NLI dataset covering 15 languages including Chinese. 

\begin{table}[t]
    \centering
\begin{CJK}{UTF8}{gbsn}
{\footnotesize
\begin{tabular}{ p{5cm} | p{1.7cm} }
\multicolumn{1}{c}{\textbf{Premise}} & \multicolumn{1}{c}{\textbf{Hypothesis}} \\ \hline
a. \textcolor{darkgray}{Louisa May Alcott和Nathaniel Hawthorne 住在Pinckney街道，而 那个被Oliver Wendell Holmes称为 “晴天街道 的Beacon Street街道住着有些喜欢自吹自擂的历史学家 William Prescott}  \newline  \emph{\textbf{Eng.:} Louisa May Alcott and Nathaniel Hawthorne lived on Pinckney street, but on Beacon Street street, which is named ``Sunny Street by Oliver Wendell Holmes, lived the bragging historian William Prescott. [sic]}
& \textcolor{darkgray}{Hawthorne住在Main Street上}  \newline \emph{Eng.: Hawthorne lived on Main Street}.
\\ \hline
b. \textcolor{darkgray}{运行 Slient，运行Deep，运行答案} \newline  \emph{Eng.: run Slient, run Deep, run answer. [sic]} &
\textcolor{darkgray}{悄悄的逃走} \newline \emph{\textbf{Eng.:} secretly escape.}
\\ \hline
\end{tabular}
}
\end{CJK}
\caption{Examples from crowd-translated XNLI development set \cite{xnli}, showing problems of \emph{translationese} (top) and poor translation quality (bottom).}
    \label{fig:first_example}
\end{table}

While automatically translated data have proven to be useful in many contexts, such as  cross-lingual representation learning \cite{siddhant2019evaluating},  there are well-known issues, especially when used in place of human annotated, quality controlled  data. One issue concerns limitations in the quality of automatic translations, resulting in incorrect or unintelligible sentences (e.g., see Table~\ref{fig:first_example}b). But even if the translations are correct, they suffer from ``translationese", resulting in unnatural language, since lexical and syntactic choices are copied from the source language even though they are untypical for the target language \citep{koppel2011ACLtranslationese,stylevar2018,translationese-chinese}. 

A related issue is that a translation approach also copies the cultural context of the source language, such as an overemphasis on Western themes or cultural situations. The latter two issues are shown in Table~\ref{fig:first_example}a, %
where many English names are directly carried over into the Chinese translation, along with aspects of English syntax, such as long relative clauses, which are common in English but dispreferred in Chinese \cite{lin2011chinese}.

\paragraph{Systems:} 
\added{As inference is closely related to logic, there has always been a line of research building logic-based or logic-and-machine-learning hybrid models for NLI/RTE problems \citep[e.g.][]{MacCartney,Abzianidze15,martinez2017,YanakaMMB18,monalog}. } 
However, in recent years, large datasets such as SNLI and MNLI have been almost exclusively approached by deep learning models. 
For examples, several transformer architectures achieve impressive results on MNLI, with current state-of-the-art T5 \cite{2019t5} reaching 92.1/91.9\% accuracy on the matched and mismatched sets. 

Re-implementations of these transformer models for Chinese have led to similar successes on related tasks. For example, \citet{cui2019pretraining} report that a large RoBERTa  model \cite{roberta}, pre-trained with whole-word masking, achieves the highest accuracy (81.2\%) among their transformer models on XNLI.
In the CLUE benchmark \cite{clue}, the same RoBERTa model also achieves the highest aggregated score from eight tasks.
We will use this model to establish baselines on our new dataset.

\paragraph{\added{Biases:}}
The advances in dataset creation have led to an increased awareness of systematic biases in existing datasets \cite{gururangan2018annotation}, as measured through \emph{partial-input baselines}, e.g., the \emph{hypothesis-only} baselines explored in \citet{poliak2018hypothesis} where a model can achieve high accuracy by only looking at the hypothesis and ignoring the premise completely \added{(see also \citet{feng2019misleading})}. These biases have been mainly associated with the annotators (crowd workers in MNLI's case) who use certain strategies to form hypotheses of a specific label, e.g., adding a negator for contradictions. 

There have been several recent attempts to reduce such biases \citep{belinkov2019don,sakaguchi2019winogrande,le2020adversarial,anli}. 
\added{There has also been a large body of work using probing datasets/tasks to stress-test NLI models trained on datasets such as SNLI and MNLI, in order to expose the weaknesses and biases in either the models or the data \citep{dasgupta2018evaluating,naik2018stress,hans}. }
For this work, we closely monitor the hypothesis-only and other biases but leave systematic filtering/bias-reduction/stress-testing for future work. \added{An interesting future challenge will involve seeing how such techniques, which focus exclusively on English, transfer to other languages such as Chinese.}

{ \renewcommand{\arraystretch}{1.5} %
\begin{table*}[t]
	\centering
	\small
	\begin{tabular}{l|p{4cm}|l|l|l|l}\toprule
		\multirow{2}{*}{Subsets} & \multirow{2}{*}{Instructions}  & \multicolumn{4}{c}{\# Pairs / Mean length of hypothesis \textit{H} in characters} \\\cmidrule{3-6}
		& &  Total & \texttt{easy} & \texttt{medium} & \texttt{hard} \\\hline
\textsc{Single} & same as MNLI; one \textit{H} per label & 11,986 / 10.9 & n.a. & n.a. & n.a. \\
\textsc{Multi} & three \textit{H}s per label & 12,328 / 10.4 & 4,836 / 9.9 & 4,621 / 10.6 & 2,871 / 11.0 \\
\textsc{MultiEncourage} & \textsc{Multi} + encouraging annotators to use fewer negators and write more diverse hypotheses & 16,584 / 12.2 & 6,263 / 11.5 & 6,092 / 12.5 & 4,229 / 12.7 \\
\textsc{MultiConstraint} & \textsc{Multi} + constraints on the negators used in contradictions & 15,627 / 12.0 & 5,668 / 11.6 & 5,599 / 12.2 & 4,360 / 12.4 \\\hline
total &  & 56,486 / 11.5 &  &  & \\
		\bottomrule
	\end{tabular}
	\caption{Information on the four subsets of data collected. 
	Premises in all subsets are drawn from the same pool of text from five genres. \texttt{easy/medium/hard} refers to the 1st/2nd/3rd hypothesis written for the same premise and inference label. Number of pairs in the \texttt{hard} condition is smaller because not all premises and all labels have a third hypothesis. See section \ref{sec:hypo:generation} for details of the subsets. \label{tab:4-batch}}
\end{table*}
} %

\section{Creating OCNLI}

Here, we describe our data collection and annotation procedures. Following the standard definition of NLI \cite{dagan-etal-2006-direct}, our data consists of ordered pairs of sentences, one \textit{premise} sentence and  one \textit{hypothesis} sentence, annotated with one of three labels: %
Entailment, Contradiction, or Neutral (see examples in Table~\ref{tab:ocnli:examples}).

Following the strategy that \citet{mnli} established for MNLI, we start by selecting a set of premises from a collection of multi-genre Chinese texts, see Section~\ref{sec:select:premise}. We then elicit hypothesis annotations based on these premises using expert annotators (Section~\ref{sec:hypo:generation}). We develop novel strategies to ensure that we elicit diverse hypotheses. %
We then describe our verification procedure in Section~\ref{sec:verification}.

\subsection{Selecting the Premises \label{sec:select:premise}}

Our premises are drawn from the following five text genres: government documents, news, literature, TV talk shows, and telephone conversations. The genres were chosen to ascertain varying degrees of formality, and they were collected from different primary Chinese sources.  The government documents are taken from annual Chinese government work reports\footnote{\urlstyle{rm}\url{http://www.gov.cn/guowuyuan/baogao.htm}, last visited 4/21/2020, same below.}. The news data are extracted from the news portion of the Lancaster Corpus of Mandarin Chinese \citep{LCMC}. The data in the literature genre are from two contemporary Chinese novels\footnote{\textit{Ground Covered with Chicken Features} by Liu Zhenyun, \textit{Song of Everlasting Sorrow} by Wang Anyi.}, and the TV talk show data and telephone conversations are extracted from transcripts of the talk show \textit{Behind the headlines with Wentao}\footnote{\urlstyle{rm}\url{http://phtv.ifeng.com/listpage/677/1/list.shtml}.} and the Chinese Callhome transcripts \citep{callhome-zh}.

As for pre-processing, annotation symbols in the Callhome transcripts were removed and we limited our premise selection to sentences containing 8 to 50 characters.

\subsection{Hypothesis Generation \label{sec:hypo:generation}}

One issue with the existing data collection strategies in MNLI is that humans tend to use the simplest strategies to create the hypotheses, such as negating a sentence to create a contradiction. This makes the problem unrealistically easy. 
To create more realistic, and thus more challenging data, we propose a new hypothesis elicitation method called \textit{multi-hypothesis} elicitation. We collect four sets of inference pairs and compare the proposed method with the MNLI annotation method, where a single annotator creates an entailed sentence, a neutral sentence and a contradictory sentence given a premise (Condition: \textsc{Single}).

\paragraph{Multi-hypothesis elicitation}
In this newly proposed setting, we ask the writer to produce \textit{three} sentences per label, resulting in three entailments, three neutrals and three contradictions for each premise (Condition: \textsc{Multi}). I.e. we obtain a total of nine hypotheses if the writer is able to come up with that many inferences, which is indeed the case for most premises in our experiment.
Our hypothesis is that by asking them to produce three sentences for each type of inference, we push them to think beyond the easiest case. 
We call the 1st, 2nd and 3rd hypothesis by an annotator per label \texttt{easy}, \texttt{medium} and \texttt{hard} respectively, with the assumption that they start with the easiest inferences and then move on to harder ones. First experiments show that \textsc{Multi} is more challenging than \textsc{Single}, and at the same time, inter-annotator agreement is slightly higher than for \textsc{Single} (see section~\ref{sec:verification}).

\begin{CJK}{UTF8}{gbsn}
However, we also found that \textsc{Multi} introduces more hypothesis-only bias. 
Especially in contradictions, negators such as 没有 (``no/not'') stood out as cues, similar to what had been reported in SNLI and MNLI \citep{poliak2018hypothesis,gururangan2018annotation,pavlick2019inherent}. Therefore we experiment with two additional strategies to control the bias, resulting in \textsc{MultiEncourage} (\textit{encourage} the annotators to write more diverse hypothesis) and \textsc{MultiConstraint} (put \textit{constraints} on what they can produce), which will be explained in detail below. 

These four strategies result in four different subsets. Table~\ref{tab:4-batch} gives a summary of these subsets. 
\end{CJK}

\paragraph{Instructions for hypothesis generation} The basis of our instructions are very similar to those for MNLI, but we modified them for each setting: 

\paragraph{\textsc{Single}} We asked the writer to produce one hypothesis per label, same as MNLI\footnote{See Appendix \ref{sec:instructions} for the complete instructions.}.

\paragraph{\textsc{Multi}} Instructions are the same except that we ask for three hypotheses per label.

\begin{table*}[t]
	\centering
	\small
	\begin{tabular}{l|rrr|rrrr} 
	\toprule
		& \bf SNLI$^{\dagger}$  & \bf MNLI$^{\dagger}$  & \bf XNLI$^{\dagger}$   & \multicolumn{4}{c}{\bf OCNLI} \\
		 &  &  & \bf   & \bf \textsc{Single} &\bf \textsc{Multi} &
		\bf \textsc{MultiEnc}  &\bf \textsc{MultiCon} \\\midrule
		\# pairs in total & 570,152 & 432,702 & 7,500 & 11,986 & 12,328 & 16,584 & 15,627 \\
		\# pairs relabeled & 56,941 & 40,000 & 7,500 & 1,919 & 1,994 & 3,000 & 3,000 \\
		\% relabeled & 10.0\% & 9.2\% & 100.0\% & 16.0\% & 16.2\% & 18.1\% & 19.2\% \\\midrule
		5 labels agree (unanimous) & 58.3\% & 58.2\% & na & 62.1\% & 63.5\% & 57.2\% & 57.6\% \\
		4+ labels agree & na & na & na & 82.2\% & 84.8\% & 82.0\% & 80.8\% \\
		3+ labels agree & \textbf{98.0\%} & \textbf{98.2\%} & \textbf{93.0\%} & \textbf{98.6}\% & \textbf{98.8\%} & \textbf{98.7\%} & \textbf{98.3\%} \\\midrule
		Individual label $=$ gold label & 89.0\% & 88.7\% & na & 88.1\% & 88.9\% & 87.0\% & 86.7\% \\
		Individual label $=$ author's label & 85.8\% & 85.2\% & na & 81.8\% & 82.3\% & 80.2\% & 79.7\% \\\midrule
		Gold label $=$ author's label & 91.2\% & 92.6\% & na & 89.8\% & 89.6\% & 89.6\% & 88.2\% \\
		Gold label $\ne$ author's label & 6.8\% & 5.6\% & na & 8.8\% & 9.2\% & 9.0\% & 10.1\% \\
		No gold label (no 3 labels match) & 2.0\% & 1.8\% & na & 1.4\% & 1.2\% & 1.3\% & 1.7\% \\
		\bottomrule
	\end{tabular}
	\caption{Results from labeling experiments for the four subsets. \textsc{MultiEnc}: \textsc{MultiEncourage}; \textsc{MultiCon}: \textsc{MultiConstraint}.  $^{\dagger}$ = numbers for SNLI, MNLI, XNLI are copied from the original papers \protect \citep{snli,mnli,xnli}. For XNLI, the numbers are for the English portion of the dataset, which is the only language that has been relabelled. \label{tab:labelling}}
\end{table*}

\paragraph{\textsc{MultiEncourage}} We \textit{encouraged}  the writers to write high-quality hypotheses by telling them explicitly which types of data we are looking for, and promised a monetary bonus to those who met our criteria after we examined their hypotheses. 
Among our criteria are: 1) we are interested in \textit{diverse} ways of making inferences, and 2) we are looking for contradictions that do \textit{not} contain a negator.

\paragraph{\textsc{MultiConstraint}} We put \textit{constraints} on hypothesis generation by specifying that \textit{only one out of the three contradictions can contain a negator}, and that we would randomly check the produced hypothesis, with violations of the constraint resulting in lower payment. We also provided extra examples in the instructions to demonstrate contradictions without negators. These examples are drawn from the hypotheses collected from prior data. 

\added{We are also aware of other potential biases or heuristics in human-elicited NLI data such as the lexical overlap heuristic \citep{hans}. Thus in all our instructions, we made explicit to the annotators that no hypothesis should overlap more than 70\% with the premise. However, examining how prevalent such heuristics are in our data requires constructing new probing datasets for Chinese, which is beyond the scope of this paper. }

\paragraph{Annotators} We hired 145 undergraduate and graduate students from several top-tier Chinese universities to produce hypotheses. 
All of the annotators (\textit{writers}) are native speakers of Chinese and are majoring in Chinese or other languages. 
 They were paid roughly 0.3 RMB (0.042 USD) per P-H pair. No single annotator produced an excessive amount of data to avoid annotator-bias \added{(for a discussion of this, see \citet{annotator2019Geva})}.

\subsection{Data Verification \label{sec:verification}} 

Following SNLI and MNLI, we perform data verification, where each premise-hypothesis pair is assigned a label by four independent annotators (\textit{labelers}). Together with the original label assigned by the annotator, each pair has five labels. We then use the majority vote as the gold label. We selected a subset of the writers from the hypothesis generation experiment to be our labelers. For each subset, about 15\% of the total data were randomly selected and relabeled. The labelers were paid 0.2 RMB (0.028 USD) for each pair.

\paragraph{Relabeling results \label{sec:relabel:results}}
Our results, shown in Table~\ref{tab:labelling}, are very close to the numbers reported for SNLI/MNLI, with labeler agreement even higher than SNLI/MNLI for \textsc{Single} and \textsc{Multi}.

Crucially, the three \textsc{Multi} subsets, created using the three variants of the \textit{multi-hypothesis} generation method, have similar agreement to MNLI, suggesting that producing nine hypotheses for a given premise is feasible. Furthermore, the agreement rates on the \texttt{medium} and \texttt{hard} portions of the subsets are only slightly lower than on the \texttt{easy} portion, with agreement rates of 3 labels at least 97.90\% (see Table~\ref{tab:labelling:3portions} in the Appendix), suggesting that our data in general is of high quality. 
Agreement is lower for \textsc{MultiConstraint}, showing that it may be difficult to produce many hypotheses under these constraints.

In a separate relabeling experiment, we examine the quality of human-translated examples from the XNLI dev set. The results show considerably lower agreement: The majority vote of our five annotators only agree with the XNLI gold-label 67\% of the time, as compared to the lowest rate of 88.2\% on \textsc{MultiConstraint}. 
Additionally, 11.6\% of the XNLI dev examples in Chinese contain more than 10 Roman alphabets, which are extremely rare in original, every-day Chinese speech/text.
These \added{results} suggest that XNLI is less suitable as validation set for Chinese NLI, and thus we excluded XNLI dev set in our evaluation. For \added{further} details, see Appendix \ref{sec:xnli:label}.

\subsection{The Resulting Corpus \label{sec:ocnli:datasplit}}

Overall, we have a corpus of more than 56,000 pairs of inference pairs in Chinese. We have randomized the total of 6,000 \textit{relabeled} pairs from \textsc{MultiEncourage} and \textsc{MultiConstraint} and used them as the development and test sets, each consisting of 3,000 examples.  
All pairs from \textsc{Single} and \textsc{Multi}, plus the remaining 26,211 pairs from \textsc{MultiEncourage} and \textsc{MultiConstraint} are used for the training set,  about 50,000 pairs\footnote{We note that given the constraints of having equal number of \texttt{easy}, \texttt{medium} and \texttt{hard} examples in dev/test sets, the resulting corpus ended up having high premise overlap between training and dev/test sets, in contrast to the original MNLI design. To ensure that such premise overlap does not bias the current models and inflate performance, we experimented with a smaller \textbf{non-overlap} train and test split, which was constructed by filtering parts of the training. This lead to comparable results, despite the non-overlap being much smaller in size, which we detail in Appendix \ref{sec:no-overlap}. Both the \textbf{overlap} and \textbf{non-overlap} splits will be released for public use, as well as part of the the public leaderboard at \urlstyle{rm}\url{https://www.cluebenchmarks.com/nli.html}. }. 
This split ensures that all labels in the development and test sets have been verified, and the number of pairs in the \texttt{easy}, \texttt{medium} and \texttt{hard} portions are roughly the same in both sets. It is also closer to a realistic setting where contradictions without negation are much more likely.
Pairs that do not receive a majority label in our relabeling experiment are marked with ``-'' as their label, and can thus be excluded if necessary. %

\section{Experimental Investigation of OCNLI}

\subsection{Experimental Setup}
To demonstrate the difficulty of our dataset, we establish baselines using several widely-used NLI models tailored to Chinese\footnote{Additional details about all of our models and hyper-parameters are included as supplementary material.}. This includes the baselines originally used in \citet{mnli} such as the continuous bag of words (CBOW) model, the biLSTM encoder model and an implementation of ESIM \cite{chen2017lstm}\footnote{We use a version of the implementations from \urlstyle{rm}\url{https://github.com/NYU-MLL/multiNLI}.}. In each case, we use Chinese character embeddings from \citet{chinese-emb} in place of the original GloVe embeddings. 

We also experiment with state-of-the-art pre-trained transformers for Chinese \cite{cui2019pretraining} using the fine-tuning approach from \citet{bert}. Specifically, we use the Chinese versions of BERT-base \cite{bert} and RoBERTa-large \cite{roberta} with whole-word masking (see details in \citet{cui2019pretraining}). In both cases, we rely on the publicly-available TensorFlow implementation provided in the CLUE benchmark \cite{clue}\footnote{See: \urlstyle{rm}\url{https://github.com/CLUEbenchmark/CLUE}}.  Following \citet{bowman2020collecting}, we also fine-tune \emph{hypothesis-only}  variants of our main models to measure annotation artifacts. 

To measure human performance, we employed an additional set of 5 Chinese native speakers to annotate a sample ($300$ examples) of our OCNLI test set. This follows exactly the strategy used in \citet{glue-human} for measuring human performance in GLUE, and provides a \emph{conservative} estimate of human performance in that annotators were provided with minimal amounts of task training (see Appendix \ref{sec:human:baseline} for details).

\paragraph{Datasets} In addition to experimenting with OCNLI, we also compare the performance of our main models against models fine-tuned on the Chinese training data of XNLI \cite{xnli} (an automatically translated version of MNLI), as well as combinations of OCNLI and XNLI. The aim of these experiments is to evaluate the relative advantage of automatically translated data. 
We also compare both models against the CLUE diagnostic test from \citet{clue}, which is a set of 514 NLI problems that was annotated by \added{an independent set of} Chinese linguists. 

To analyze the effect of our different hypothesis elicitation strategies, we look at model performance on different subsets of OCNLI. Due to the way in which the data is partitioned (all of \textsc{Single} and \textsc{Multi} are in the training set), it is difficult to fine-tune on OCNLI and test on all four subsets. We instead use an XNLI trained model, which is independent of any biases related to our annotation process, to probe the difficulty of our different subsets.

\begin{table*}[t]
\centering
\begin{tabular}{l|lll|ll}\toprule
Maj. &  CBOW & biLSTM & ESIM & BERT & RoBERTa \\\midrule
 38.1 & 55.7 \fz{(0.5)} & 59.2 \fz{(0.5)} & 59.8 \fz{(0.4)} & 72.2 \fz{(0.7)} & 78.2 \fz{(0.7)} \\ \bottomrule
\end{tabular}
\caption{Test performance on OCNLI for all baseline models. Majority label is \textit{neutral}. \added{We report the mean accuracy \% across five training runs with random re-starts (the standard deviation is shown in parentheses).} \label{tab:all}}
\end{table*}

\subsection{Baseline Results and Analysis}

In this section, we describe our main results. 

\begin{table*}[t]
\centering

\begin{tabular}{ll|rr|rr|c}
\toprule
\multicolumn{2}{c|}{\textbf{Fine-tuning data} / size} & \multicolumn{2}{c|}{\textbf{OCNLI} / 50k} & \multicolumn{2}{c|}{\textbf{XNLI} / 392k} & \multicolumn{1}{c}{\textbf{Combined} / 443k} \\
Test data & size & BERT & RoBERTa & BERT & RoBERTa & RoBERTa \\ \midrule
OCNLI human & 300 & \multicolumn{2}{c|}{90.3* (OCNLI.test)} & &  &  \\
OCNLI.dev & 3k & 74.5 \fz{(0.3)} & \textbf{78.8} \fz{(1.0)} & 66.8 \fz{(0.5)} & 70.5 \fz{(1.0)} &  76.4 \fz{(1.3)} \\
OCNLI.test & 3k & 72.2 \fz{(0.7)} & \textbf{78.2} \fz{(0.7)} & 66.7 \fz{(0.3)} & 70.4 \fz{(1.2)} &  75.6 \fz{(1.2)} \\
CLUE diagnostics & 0.5k & 54.4 \fz{(0.9)} & 61.3 \fz{(1.3)} & 53.0 \fz{(0.9)} & 62.5 \fz{(2.9)} &  \textbf{63.7} \fz{(2.4)} \\ \bottomrule
\end{tabular}
\caption{Accuracy on OCNLI, finetuned on OCNLI, XNLI and Combined (50k OCNLI combined with 392k XNLI).
*: See Appendix \ref{sec:human:baseline} for details about the human baseline. \added{As in Table~\ref{tab:all}, we report the mean accuracy \% across five training runs with the standard deviation shown in parenthesis.}
}
\label{tab:res:1}
\end{table*}

\paragraph{How Difficult is OCNLI?} To investigate this, we train/fine-tune all five neural architectures on OCNLI training data and test on the OCNLI test set. The main results %
are shown in Table~\ref{tab:all}. All of the non-transformer models perform poorly while BERT and RoBERTa reach a $\sim$20 percentage-point advantage over the strongest of these models (ESIM). This shows the relative strength of pre-trained models on our task. 

We find that while transformers strongly outperform other baseline models, our best model, based on RoBERTa, is still about 12 points below human performance on our test data (i.e., 90.3\% versus 78.2\%). This suggests that models have considerable room for improvement, and provides additional evidence of task difficulty. In comparison, these transformer models reach human-like performance in many of the GLUE \cite{glue} and SuperGLUE \cite{superglue} tasks.  
For NLI specifically, the performance of the English RoBERTa on MNLI is 90.4\%, and only about 2 percentage-points below the human score \citep{bowman2020collecting,glue-human}. 
We see a similar trend for BERT, which is about 18 points behind human performance on OCNLI, but the difference is roughly 8 points for MNLI \cite{bert}. We also see much room for improvement on the CLUE diagnostic task, where our best model achieves only 61.3\% (a slight improvement over the result reported in \citet{clue}).

\begin{table}[t]
\centering
\begin{tabular}{l|ll} \toprule
Test data & BERT & RoBERTa \\\midrule
OCNLI\_dev & 65.3 & 65.7 \\
OCNLI\_test & 64.3 & 65.0 \\
OCNLI\_test\_easy & 63.5 & 64.0 \\
OCNLI\_test\_medium & 63.9 & 65.6 \\
OCNLI\_test\_hard & 65.5 & 65.5 \\\midrule
MNLI & na. & 62.0 \\\bottomrule
\end{tabular}
\caption{Hypothesis-only baselines for OCNLI (fine-tuned on OCNLI.train) and MNLI (retrieved from \protect \citet{bowman2020collecting}). \label{tab:res:hypo:only} }
\end{table}

We also looked at how OCNLI fares on hypothesis-only tests, where all premises in train and test are replaced by the same non-word, thus forcing the system to make predictions on the hypothesis only. 
Table~\ref{tab:res:hypo:only} shows the performance of these models on different portions of OCNLI. These results show that our elicitation gives rise to annotation artifacts in a way similar to most benchmark NLI datasets (e.g., OCNLI: $\sim66\%$; MNLI  $\sim62\%$ and SNLI: $\sim69\%$, as reported in \citet{bowman2020collecting} and \citet{poliak2018hypothesis}, respectively). We specifically found that negative polarity items (``any'', ``ever''),  negators and ``only'' are among the indicators for contradictions, whereas ``at least'' biases towards entailments. We see no negators for the \textsc{MultiConstraint} subset, which shows the effect of putting constraints on the hypotheses that the annotators can produce. Instead, ``only'' is correlated with contradictions. A more detailed list is shown in Figure~\ref{tab:bias:pmi}, listing individual word and label pairs with high pairwise mutual information (PMI). PMI  was also used by \citet{bowman2020collecting} for the English NLI datasets. 

Given the large literature on adversarial filtering \cite{le2020adversarial} and adversarial learning \cite{belinkov2019don} for NLI, which have so far been limited to English and on much larger datasets that  are easier to filter, we see extending these methods to our dataset and Chinese as an interesting challenge for future research.

\begin{CJK}{UTF8}{gbsn}
\begin{table}[t]
\centering\small
\begin{tabular}{llll}\toprule
Word & Label & PMI & Counts \\\midrule
OCNLI &  &  &  \\
任何 \textit{any} & contradiction & 1.02 & 439/472 \\
从来 \textit{ever} & contradiction & 0.99 & 229/244 \\
至少 \textit{at least}  & entailment & 0.92 & 225/254 \\
\midrule
\textsc{Single} &  &  &  \\
任何 \textit{any} & contradiction & 0.89 & 87/90 \\
没有 \textit{no} & contradiction & 0.83 & 582/750 \\
无关 \textit{not related} & contradiction & 0.72 & 39/42 \\
\midrule
\textsc{Multi} &  &  &  \\
任何 \textit{any} & contradiction & 0.92 & 97/103 \\
没有 \textit{no} & contradiction & 0.88 & 721/912 \\
从来 \textit{ever} & contradiction & 0.75 & 42/46 \\
\midrule
\textsc{MultiEncourage} &  &  &  \\
任何 \textit{any} & contradiction & 0.98 & 198/212 \\
从来 \textit{ever} & contradiction & 0.96 & 131/137 \\
至少 \textit{at least} & entailment & 0.82 & 81/91 \\
\midrule
\textsc{MultiConstraint} &  &  &  \\
至少 \textit{at least} & entailment & 0.91 & 105/110 \\
只有 \textit{only} & contradiction & 0.86 & 179/216 \\
只 \textit{only} & contradiction & 0.77 & 207/280 \\
\bottomrule
\end{tabular}
\caption{Top 3 \texttt{(word, label)} pairs according to PMI for different subsets of OCNLI. \label{tab:bias:pmi}}
\end{table}
\end{CJK}

\paragraph{Comparison with XNLI} To ensure that our dataset is not easily solved by simply training on existing translations of MNLI, we show the performance  of BERT and RoBERTa when trained on XNLI but tested on OCNLI. The results in Table~\ref{tab:res:1} (column XNLI) show a much lower performance than when the systems are trained on OCNLI, even though XNLI contains 8 times more examples.\footnote{To ensure that this result is not unique to XNLI, we ran the same experiments using CMNLI, which is an alternative translation of MNLI used in CLUE, and found comparable results.} While these results are not altogether comparable, given that the OCNLI training data was generated from the same data sources and annotated by the same annotators (see \citet{annotator2019Geva}), we still see these results as noteworthy given that XNLI is currently the largest available multi-genre NLI dataset for Chinese. The results are indicative of the limitations of current models trained solely on translated data. More strikingly, we find that when OCNLI and XNLI are combined for fine-tuning (column Combined in Table~\ref{tab:res:1}), this  improves performance over the results using XNLI, but reaches lower accuracies than fine-tuning on the considerably smaller OCNLI (except for the diagnostics).

\begin{figure}[t]
    \centering
    \includegraphics[width=0.48\textwidth]{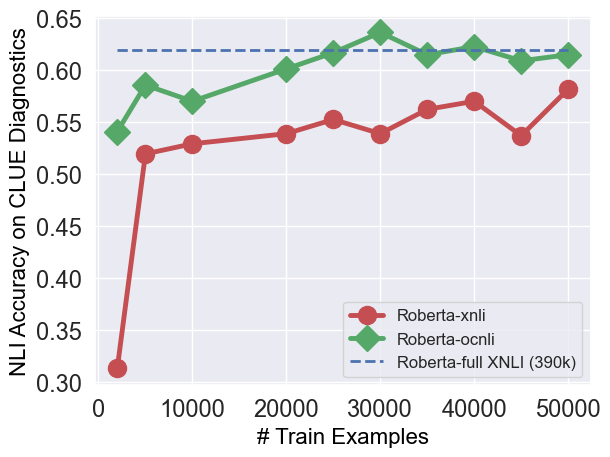}
    \caption{Ablation over the number of fine-tuning examples for RoBERTa fine-tuned on OCNLI vs. XNLI.} %
    \label{fig:curve}
\end{figure}

Figure~\ref{fig:curve} shows a learning curve comparing model performance on the independent CLUE diagnostic test. Here we see that the OCLNI model reaches its highest performance at 30,000 examples while the XNLI model still shows improvements on 50,000 examples. Additionally, OCNLI reaches the same performance as the model finetuned on the full XNLI set, at around 25,000 examples.
This provides additional evidence of the  importance of having reliable human annotation for NLI data.

\begin{table}[t]
\centering\small

\resizebox{0.95\textwidth}{!}{\begin{minipage}{\textwidth}
\begin{tabular}{lrrrr}\toprule
 & \textsc{Single} & \textsc{Multi}  & \textsc{MultiEnc} & \textsc{MultiCon}  \\ \midrule
\multicolumn{5}{l}{BERT: fine-tune on XNLI}\\\midrule
dev\_full & 77.3 & 73.6 & 68.6 & 65.8 \\
easy & na. & 74.0 & 70.1 & 68.4 \\
medium &na. & 74.3 & 69.6 & 65.9 \\
hard & na.& 72.5 & 66.2 & 63.1 \\
\midrule
\multicolumn{5}{l}{RoBERTa: fine-tune on XNLI} \\\midrule
dev\_full & 78.9 & 77.3 & 71.3 & 70.8 \\
easy & na. & 77.2 & 72.8 & 73.5 \\
medium & na.& 78.6 & 71.7 & 70.2 \\
hard & na. & 76.2 & 69.4 & 68.7 \\
\bottomrule
\end{tabular} %
\end{minipage}}
\caption{Accuracy of XNLI-finetuned models, tested on relabelled parts of different OCNLI subsets. 
\label{tab:res:4batches} }
\end{table}

\paragraph{Understanding the OCNLI Subsets} To better understand the effect of having  three annotator hypotheses per premise, constituting three difficulty levels, and having four elicitation modes, we carried out a set of experiments  with XNLI-finetuned models on the different subsets. We used XNLI to avoid imposing specific preferences on the models.  Table~\ref{tab:res:4batches} shows  a consistent decrease in accuracy from \textsc{Single} through \textsc{MultiConstraint}, and a mostly consistent decrease from easy to hard (exception: between easy and medium in \textsc{Multi}). Both trends 
suggest that \textit{multi-hypothesis} elicitation and improved instructions lead to more challenging elicited data.

\section{Conclusion}
In this paper, we presented the Original Chinese Natural Language Inference (OCNLI) corpus, the first large-scale, non-translated NLI dataset for Chinese. Our dataset is composed of 56,000 premise-hypothesis pairs, manually created by university students with a background in language studies, using  premises from five genres and  an enhanced protocol \added{from the original MNLI annotation scheme}. Results using BERT and RoBERTa show that our dataset is challenging for the current best pre-trained transformer models, the best of which is $\sim12$ percentage-points below human performance. \added{We also demonstrate the relative advantage of using our human constructed dataset over machine translated NLI such as XNLI.} \added{To encourage more progress on Chinese NLU, we are making our dataset publicly available for the research community at \url{https://github.com/CLUEbenchmark/OCNLI} and will be hosting a leaderboard in the Chinese Natural Language Understanding (CLUE) \cite{clue} benchmark (\url{https://www.cluebenchmarks.com/nli.html}). }

\added{Given the wide impact that large-scale NLI datasets, such as SNLI and MNLI, have had on recent progress in NLU for English, we hope that our resource will likewise help accelerate progress on Chinese NLU. In addition to making more progress on Chinese NLI, future work will also focus on using our dataset for doing Chinese model probing (e.g., building on work such as \citet{warstadt2019investigating,probing2020,jeretic2020natural}) and sentence representation learning \cite{reimers2019sentence}, as well as for investigating bias-reduction techniques \cite{clark2019don,belinkov2019don,le2020adversarial} for languages other than English.}

\section*{Acknowledgements}

We thank all our annotators without whom this work wouldn't have been possible, and also Ruoze Huang, Jueyan Wu, Zhaohong Wu and Xiaojie Gong for their help in the annotation process. We are grateful for the suggestions from our anonymous reviewers and the CL colloquium at Indiana University. This work was supported by the CLUE benchmark and the Grant-in-Aid of Doctoral Research from Indiana University Graduate School.  \added{Special thanks to the beaker team at AI2 for providing technical support for the beaker experiment platform. Computations on \url{beaker.org} were supported in part by credits from Google Cloud.}

\bibliographystyle{acl_natbib}
\bibliography{main}

\appendix

\section{Instructions for Hypothesis Generation\label{sec:instructions}}

(\textit{the instructions are originally in Chinese; translated to English for this paper})

~

Welcome to our sentence writing experiment. Our aim is to collect data for making inferences in Chinese. 
In this experiment, you will see a sentence (A), which describes an event or a scenario, for example:

\begin{mdframed}
Sentence A: \\
\textbf{John won the first prize in his company's swimming competition last year. }
\end{mdframed}

You task is to write three types of sentences based on the information in sentence A, as well as your common sense. 

\begin{itemize}
	\item Type 1: a sentence that is definitely true, based on the information in sentence A, e.g.,
	\begin{itemize}
		\item John can swim
		\item John won a prize last year
		\item John's company held a swimming competition last year
	\end{itemize}
\item Type 2: a sentence that might be true (but might also be false), based on the information in sentence A, e.g.,
\begin{itemize}
	\item John's company held the swimming competition last March
	\item Tom ranked second in last year's swimming competition
	\item John can do the butterfly style
\end{itemize}
\item Type 3: a sentence that cannot be true, based on the information in sentence A, e.g.,
\begin{itemize}
	\item John has not swum before
	\item John did not get any prize from the company's swimming competition last year
	\item John's company only hold table tennis competitions
\end{itemize}
\end{itemize}

You will see 50 sentence A. For each sentence A, you need to write three sentences, one for each type. In total 
you will write 150 sentences. If there is a problem with sentence A, please mark it as ``x''. Please refer to FAQ for more examples and further details of the task.

\section{Relabeling Results for Different Portions}

In Table \ref{tab:labelling:3portions}, we present labeler agreement for different portions of \textsc{Multi}, \textsc{MultiEncourage} and \textsc{MultiConstraint}. We observe that the \texttt{medium} and \texttt{hard} portions in general have lower inter-annotator agreement, but still comparable to SNLI and MNLI. This suggests that writing three hypothese for each label is a feasible and reliable strategy.

\begin{table*}[t]
	\begin{centering}
	\scalebox{0.95}{
		\footnotesize
		\begin{tabular}{l|lll|lll|lll}
			\toprule
			& \multicolumn{3}{c}{\bf \textsc{Multi}} &\multicolumn{3}{c}{\bf \textsc{MultiEncourage}} &\multicolumn{3}{c}{\bf \textsc{MultiConstraint}}  \\
			\bf Statistic & easy & medium & hard &  easy &   medium &  hard &  easy &  medium &  hard \\\midrule
            \# pairs relabelled & 668 & 664 & 662 & 1,002 & 999 & 999 & 1,002 & 999 & 999 \\\midrule
            5 labels agree (unanimous) & 66.5\% & 61.4\% & 62.5\% & 58.0\% & 56.5\% & 57.2\% & 60.8\% & 57.2\% & 54.9\% \\
            4+ labels agree & 87.0\% & 82.1\% & 85.2\% & 82.2\% & 82.6\% & 81.2\% & 84.5\% & 78.6\% & 79.4\% \\
            3+ labels agree & \textbf{99.1}\% & \textbf{99.1}\% & \textbf{98.2}\% & \textbf{98.5}\% & \textbf{99.1}\% & \textbf{98.4}\% & \textbf{98.0}\% & \textbf{99.0}\% & \textbf{97.9}\% \\\midrule
            Indiv.~label $=$ gold label & 90.1\% & 88.2\% & 88.5\% & 87.1\% & 87.3\% & 86.7\% & 87.9\% & 86.5\% & 85.6\% \\
            Indiv.~label $=$ author's label & 84.5\% & 80.0\% & 82.4\% & 80.8\% & 80.8\% & 78.9\% & 82.2\% & 79.2\% & 77.6\% \\\midrule
            Gold label $=$ author's label & 91.5\% & 88.1\% & 89.3\% & 90.4\% & 91.4\% & 87.1\% & 90.1\% & 88.3\% & 86.1\% \\
            Gold label $\ne$ author's label & 7.6\% & 11.0\% & 8.9\% & 8.1\% & 7.7\% & 11.3\% & 7.9\% & 10.7\% & 11.8\% \\
            No gold label & 0.9\% & 0.9\% & 1.8\% & 1.5\% & 0.9\% & 1.6\% & 2.0\% & 1.0\% & 2.1\% \\
            \%n\_unrelated labels & 0.2\% & 0.2\% & 0.4\% & 0.2\% & 0.6\% & 0.3\% & 0.1\% & 0.1\% & 0.4\%\\
			\bottomrule
		\end{tabular}
		}
		\caption{Labeling results for different portions of \textsc{Multi}, \textsc{MultiEncourage} and \textsc{MultiConstraint}. \label{tab:labelling:3portions}}
	\end{centering}
\end{table*}

\section{Relabeling Results for XNLI Development Set \label{sec:xnli:label}}

For this experiment, we follow the same procedure as the relabeling experiment for OCNLI data. We randomly selected 200 examples from XNLI dev, and mixed them with 200 examples from our \textsc{Single} (which has already been relabelled) for another group of annotators to label. The labelers for these 400 pairs were undergraduate students who did \textit{not} participated in hypothesis generation so as to avoid biasing towards our data. 

The labeling results for XNLI are presented in Table \ref{tab:xnli:label}. 
Only 67\% of the 200 pairs have the same label from our annotators and the label given in XNLI dev. 8.5\% of the pairs are considered to be irrelevant by the majority of our annotators. 
As we mentioned in the introduction, there are other issues with XNLI such as the 
existence of many Roman alphabets (867 (11.56\%) examples in XNLI dev have more than 10 Roman alphabets) which prevent 
us from using it as proper evaluation data for Chinese NLI.

\begin{table}[t]
	\centering
	\scalebox{.9}{
	\small
	\begin{tabular}{l|ll}
	\toprule
		\bf Statistic & \bf XNLI dev & \bf \textsc{Single} \\\midrule
		\# pairs relabelled (i.e., validated) & 200 & 200 \\\midrule
		majority label $=$ \textit{original} label & \textbf{67.0}\% & \textbf{84.0}\% \\\midrule
		5 labels agree (excl.~``unrelated'') & 38.5\% & 57.5\% \\
		4+ labels agree (excl.~``unrelated'') & 57.5\% & 83.5\% \\
		3+ labels agree (excl.~``unrelated'') & \textbf{86.0}\% & \textbf{98.0}\% \\\midrule
		5 labels agree & 41.0\% & 57.5\% \\
		4+labels agree & 62.0\% & 83.5\% \\
		3+ labels agree & 94.5\% & 98.0\% \\\midrule
		majority label = ``unrelated'' & 8.5\% & 0\% \\
		\# individual ``unrelated'' labels & 125 & 11 \\
		\# incomprehensible note & 22 & 4 \\\bottomrule
	\end{tabular}
	}
\caption{Results for labeling a mixture of 200 pairs of XNLI dev Chinese and 200 pairs of \textsc{Single}, by labelers who did not participated in the hypothesis generation experiment. 
Note the XNLI dev is translated by crowd translators \protect \citep{xnli}, not MT systems.  The \textit{original} label for XNLI dev Chinese comes with XNLI, which is the same for all 15 languages. The \textit{original} label for \textsc{Single} comes from our relabeling experiments. \label{tab:xnli:label} }
\end{table}

\section{Model Details and Hyper-parameters\label{sec:models}}
We experimented with the following models:
\begin{itemize}
    \item Continuous bag-of-words (CBOW), where sentences are represented as the sum of its Chinese character embeddings, which are passed on to a 3-layer MLP.
    \item Bi-directional LSTM (biLSTM), where the sentences are represented as the average of the states of a bidirectional LSTM. 
    \item Enhanced Sequential Inference Model (ESIM), which is MNLI's implementation of the ESIM model \citep{chen2017lstm}.
    \item BERT base for Chinese (BERT), which is a 12-layer transformer model with a hidden size of 768, pre-trained with 0.4 billion tokens of the Chinese Wikipedia dump \citep{bert}. We use the implementation from the CLUE benchmark \citep{clue}\footnote{\urlstyle{rm}\url{https://www.cluebenchmarks.com/}}.
    \item RoBERTa large pre-trained with whole word masking (wwm) and extended (ext) data (RoBERTa), which is based on RoBERTa \citep{roberta} and has 24 layers with a hidden size of 1024, pre-trained with 5.4 billion tokens, released in \cite{cui2019pretraining}. We use the implementation from the CLUE benchmark.
\end{itemize}

For CBOW, biLSTM and ESIM, we use Chinese character embeddings from \citet{chinese-emb}\footnote{\urlstyle{rm}\url{https://github.com/Embedding/Chinese-Word-Vectors}}, and modify the implementation from MNLI\footnote{\urlstyle{rm}\urlstyle{rm}\url{https://github.com/NYU-MLL/multiNLI}} to work with Chinese. 

Our BERT and RoBERTa models are both fine-tuned with 3 epochs, a learning rate of 2e-5, and a batch size of 32.
Our hyper-parameters deviate slightly from those used in CLUE and \citep{cui2019pretraining}\footnote{\urlstyle{rm}\url{https://github.com/ymcui/Chinese-BERT-wwm/}}, because we found them to be better when tuned against our dev sets (as opposed to XNLI or the machine translated CMNLI in CLUE).

\section{Determining Human Baselines \label{sec:human:baseline}}

We follow procedures in \citet{glue-human} to obtain \emph{conservative} human baselines on OCNLI. Specifically, we first prepared 20 training examples from OCNLI.train and instructions similar to those in the relabeling experiment. Then we asked 5 undergraduate students who did \textit{not} participate in any part of our previous experiment to perform the labeling. They were first provided with the instructions as well as the 20 training examples, which they were asked to label after reading the instructions. Then they were given the answers and explanations of the training examples. Finally, they were given a random sample of 300 examples from the OCNLI test set  for labeling. We computed the majority label from them, and compare that against the gold label in OCNLI.test to obtain the accuracy. For pairs with no majority label, we use the most frequent label from OCNLI.test (neutral), following \citet{glue-human}. We have only 2 (0.7\%) such cases. The results are presented in Table \ref{tab:human:score}.

We performed the same experiment with 5 linguistics PhDs, who are already familiar with the NLI task from their research, and thus their results may be biased. We see a higher 5-label agreement and similar accuracy compared against the gold label of OCNLI.test. We use the score from undergraduate students as our human baseline as it is the ``unbiased'' score obtained using the same procedure as \citet{glue-human}.

The human score of OCNLI is similar to that of MNLI (92.0\%/92.8\% for match and mismatch respectively). 

\vspace{.5em}

\begin{table}[t]
\centering
{\small
\begin{tabular}{l|rr}\toprule
    annotator & undergrad & linguistics PhD \\\midrule
    \# pairs anno. & 300 & 300 \\\midrule
    accuracy (agree  &  &   \\
    w/ OCNLI.test) & \textbf{90.3}  & \textbf{89.3} \\\midrule
    5-label agree & 55.3 & 60.6 \\
    4-label agree & 82.0 & 83.3 \\
    3-label agree & 99.3 & 99.0 \\
    no majority & 0.7 & 1.0 \\\bottomrule
\end{tabular}
}
    \caption{Human score for OCNLI \label{tab:human:score}}
\end{table}

\section{More Examples from OCNLI}
We present more OCNLI pairs in Table \ref{tab:ocnli:more:examples}.

{ \renewcommand{\arraystretch}{2} %

\begin{CJK}{UTF8}{gbsn}
\begin{table*}[h]
\scalebox{.95}{
{\footnotesize
\begin{tabular}{p{6cm}p{1.3cm}p{2cm}p{5cm}} \toprule
Premise & Genre\newline \texttt{Level} & \textbf{Majority label} \newline All labels & Hypothesis \\ \midrule
是，你看他讲这个很有意思\newline  Yes, look, what he talked about is very interesting. & TV\newline hard & \textbf{Entailment}\newline E E N E E & 他讲的这个引起了我的关注 \newline What he talked about has caught my attention. \\
要根治拖欠农民工工资问题，抓紧制定专门行政法规，确保付出辛劳和汗水的农民工按时拿到应有的报酬\newline  (We need to) solve the problem of delaying wages for the migrant workers at its root and act promptly to lay out specific administrative regulations to ensure those hardworking migrant workers receive the wages they deserve in a timely manner. & GOV\newline \texttt{easy} & \textbf{Neutral}\newline N E N N N & 专门行政法规是解决拖欠工资问题的根本途径\newline  (Designing) specific administrative regulations is the most fundamental way of solving the issue of wage delays. \\
你要回来啦,就住在我这老房.\newline  If you are back, you can stay in my old house. & PHONE\newline \texttt{hard} & \textbf{Contradiction}\newline C C C C C & 我没房\newline  I don't have a house. \\
十月底回去,十一月份底回来.\newline  Going there at the end of October, be back at the end of November. & PHONE\newline \texttt{medium} & \textbf{Contradiction}\newline C C C C C & 要在那边呆两个月才回来。\newline  Will stay there for two months before coming back. \\
呃,对,我大概有,这.\newline  Er, yes, I may have (it), here. & PHONE\newline \texttt{hard} & Neutral\newline N N N N N & 是别人想问我借这个东西 \newline Someone else is trying to borrow this from me. \\
桥一顶一顶地从船上过去，好像进了一扇一扇的门 \newline Bridge after bridge was passed above the boat, just like going through door after door. & LIT\newline \texttt{medium} & \textbf{Entailment}\newline E E E E E & 有不止一座桥\newline There is more than one bridge. \\
此间舆论界普遍认为，这次自民党在众议院议席减少无疑，问题在于减少多少\newline  It  is generally believed by the media that the Liberal Democratic Party are going to lose their seats. The problem is how many. & NEWS\newline \texttt{medium} & \textbf{Contradiction}\newline C C C N N & 舆论界普遍认为，这次自民党要被驱逐出众议院。\newline  It is generally believed by the media that the Liberal Democratic Party will be ousted from the House of Representatives. \\ \bottomrule
\end{tabular}
}
}
\caption{More examples from OCNLI. \label{tab:ocnli:more:examples}}
\end{table*}
\end{CJK}
} %

\section{Filtering training data \label{sec:no-overlap}}

To mimic the MNLI setting where the training data and the evaluation data (dev/test) have no overlapping premises, we filtered out the pairs in the current training set whose premise can also be found in evaluation. This means the removal of about 20k pairs in OCNLI.train, and results in a new training set which we call OCNLI.train.small, while the development and test sets remain the same. We fine-tune the biLSTM, BERT and RoBERTa models on the new, filtered training data, and the results are presented in Table \ref{tab:train:small}. 

We observe that our models have a 1.5-2.5\% drop in performance when trained with the filtered training data. Note that OCNLI.train.small is only 60\% of OCNLI.train in size, so we consider this drop to be moderate and expected, and more likely the result of reduced training data, rather than the removal of overlapping premises.  

We will release both training sets publicly and also in our public leaderboard (\urlstyle{rm}\url{https://www.cluebenchmarks.com/nli.html}).  We note that similar strategies for controlling dataset size have been used for \textsc{WinoGrande} project \citep{sakaguchi2019winogrande} and their leaderboard (\url{https://leaderboard.allenai.org/winogrande/submissions/public}). 

\begin{table*}[t]
    \centering
    \begin{tabular}{r|lll|lll}\toprule
       Train data / size  & \multicolumn{3}{|c}{\bf OCNLI.train / 50k } & \multicolumn{3}{|c}{\bf OCNLI.train.small / 30k} \\
         & biLSTM & BERT & RoBERTa & biLSTM & BERT & RoBERTa \\ \midrule
        OCNLI.dev & 60.5 \fz{(0.4)} & 74.5 \fz{(0.3)} & 78.8 \fz{(1.0)} & 58.7 \fz{(0.3)} & 72.6 \fz{(0.9)} & 77.4 \fz{(1.0)}  \\
        OCNLI.test & 59.2 \fz{(0.5)} & 72.2 \fz{(0.7)} & 78.2 \fz{(0.7)} & 57.0 \fz{(0.9)} & 70.3 \fz{(0.9)} & 76.4 \fz{(1.2)} \\
        \bottomrule
    \end{tabular}
    \caption{Comparison of models performance fine-tuned on OCNLI.train and OCNLI.train.small. As before, we report the mean accuracy \% across five training runs with the standard deviation shown in parenthesis. \label{tab:train:small}}
\end{table*}

\end{document}